\documentclass[sn-apa]{sn-jnl}% APA Reference Style

\usepackage[FIGTOPCAP]{subfigure}
\usepackage{multirow}
\usepackage{lscape}
\usepackage{tabularx}
\usepackage{rotating}
\usepackage{tabu}
\usepackage{longtable}[=v4.13]
\usepackage{booktabs}
\usepackage{apacite}
\usepackage{amsmath}
\usepackage{enumitem}
\usepackage{algpseudocode}
\usepackage{hyperref}

\usepackage{algorithm}

\jyear{2021}%

\begin{document}

\title[Automated Imbalanced Learning]{Automated Imbalanced Learning}


\author*[1]{\fnm{Prabhant} \sur{Singh}}\email{p.singh@tue.nl}

\author[1]{\fnm{Joaquin} \sur{Vanschoren}}\email{j.vanschoren@tue.nl}

\affil[1]{\orgdiv{Department of Computer Science \& Mathematics}, \orgname{Eindhoven University of Technology}, \orgaddress{\street{Groene Loper 5}, \city{Eindhoven}, \postcode{5600MB}, \country{The Netherlands}}}


\abstract{Automated Machine Learning has grown very successful in automating the time-consuming, iterative tasks of machine learning model development. However, current methods struggle when the data is imbalanced. Since many real-world datasets are naturally imbalanced, and improper handling of this issue can lead to quite useless models, this issue should be handled carefully. This paper first introduces a new benchmark to study how different AutoML methods are affected by label imbalance. Second, we propose strategies to better deal with imbalance and integrate them into an existing AutoML framework. Finally, we present a systematic study which evaluates the impact of these  strategies and find that their inclusion in AutoML systems significantly increases their robustness against label imbalance.}




\maketitle

\section{Introduction}\label{sec1}

Many real-world datasets come with intrinsic imperfections that significantly affect the performance of machine learning models. Such datasets require a significant amount of preprocessing, and often have an uneven distribution of target classes and features. Class imbalance is a particularly challenging problem, as uneven class distributions can cause models to underperform on minority classes. 

A huge body of work has been devoted to imbalanced learning which provided us with a wide range of preprocessing techniques and models which can handle class imbalance. This growing body of work also poses a challenge: one needs to to select the algorithms and preprocessors most suitable for the datasets at hand. For example, there are more than 85 variants of SMOTE \citep{chawla2002smote, smote-variants}, and many more techniques exist \citep{survey}. The suitability of any balancing technique depends on the exact properties of the data and the other preprocessing techniques and models used in machine learning pipelines, making this a daunting task indeed.

We summarise the contributions of this work as follows:
\begin{enumerate}
    \item We propose four novel benchmarks for imbalanced learning tasks with different levels of class imbalance to better analyse how different AutoML methods behave on such problem.
    \item We present AutoBalance, an open source AutoML framework that incorporates balancing strategies as part of the AutoML process.
    \item We present a systematic study which evaluates the impact of these  strategies and find that their inclusion in AutoML systems significantly increases their robustness against label imbalance.
\end{enumerate}

In the remainder of this paper, we first discuss related work in Section \ref{relatedwork}. Section \ref{problem} formally defines the problem, and Section \ref{AutoBalance} details our solution. We introduce new benchmarks for AutoML on imbalanced data in Section \ref{benchmarks}. Section \ref{experiments} describes our experiments, which are discussed in Section \ref{analysis}. Conclusions and future work are discussed in Section \ref{conclusion}.

\section{Related Work}\label{relatedwork}
In this section we summarize the key relevant methods in the field of Imbalanced Learning, and prior work on including them in AutoML systems. 

\subsection{Imbalanced Learning}\label{imbalancedlearning}
Imbalanced Learning or Imbalanced domain learning (IDL) is one of the major problems in machine learning when applying it to a real-world setting. Imbalanced learning comes with its own set of challenges and solutions \citep{Krawczyk2016LearningFI}. The most common methods to deal with imbalanced learning problems are data preprocessing techniques and ensemble based methods. 
%We give an overview of these techniques in the sections below. 
\subsubsection{Data Preprocessing techniques}\label{datapreprocessing}
Three popular techniques are used for data preprocessing in imbalanced learning problems. Oversampling, undersampling, and under-oversampling. We list some of the popular techniques and their definitions below:

\begin{itemize}
    \item Oversampling: Oversampling replicates the data points of the minority class(es) to increase their impact on training the model. Some methods create new, synthetic data points, like the “Synthetic Minority Oversampling Technique” (SMOTE) method \citep{chawla2002smote}. SMOTE is one of the most widely used oversampling techniques with different variations, for example: SVMSMOTE, SMOTENC, ADASYN, BorderlineSMOTE and KMeansSMOTE. \cite{Kovcs2019AnEC} presents a large scale study of 85 SMOTE samplers on a number of datasets. Other oversampling techniques uses generative models, such as generative adversarial networks \citep{RodriguezBertorello2019SMateSM, Mullick2019GenerativeAM} and variational autoencoders \citep{Dai2019GenerativeOW}.
    \item Undersampling: Undersampling refers to reducing the samples from the majority classes to balance the data. A number of undersampling techniques use neighbourhood based approaches  \citep{Vuttipittayamongkol2020NeighbourhoodbasedUA}. As with oversampling techniques, there are numerous undersampling techniques, for example: Condensed Nearest Neighbour \citep{CNN}, Edited Nearest Neighbour \citep{ENN}, Instance Hardness Threshold \citep{Smith2013AnIL} and TomekLinks \citep{Tomek1976TwoMO}. There have been recent advances in undersampling  via radial based methods \citep{KOZIARSKI2020107262}.
    \item Under-oversampling: There exist a few techniques that combine both under and oversampling to make the dataset balanced. SMOTEENN \citep{Batista2004ASO} and SMOTETomek \citep{Batista2004ASO} are a few examples of these kinds of techniques. Some recent approaches use neural networks for under-oversampling \citep{Koziarski2021CSMOUTECS}
\end{itemize}

\noindent There have been number of surveys \citep{survey, survey2, survey3, book} and empirical studies \citep{Kovcs2019AnEC} which compare various sampling techniques with each other. In all such studies, the most appropriate technique depends on the dataset at hand. As such, AutoML techniques that can tune the approach to each specific dataset would be very useful. Moreover, \cite{survey} indicate that there is a strong need for better benchmarks. We aim to address both issues in this work. 

\subsubsection{Ensemble Methods}\label{subsubsec2}
Ensemble methods are another solution to tackle the problem of imbalanced datasets. Bagging \citep{Breiman2004BaggingP} and boosting \citep{boosting, Maclin1997AnEE} approaches are commonly used in imbalanced learning problems. There are also a few methods combining ensemble-based methods with sampling techniques like SMOTEBoost \citep{Chawla2003SMOTEBoostIP} and RAMOBoost \citep{Chen2010RAMOBoostRM}. Some recent approaches in using ensemble methods on imbalanced learning tasks include MESA \citep{Liu2020MESABE} which adaptively resamples the training set in iterations to get multiple classifiers and forms a cascade ensemble model. 

\subsection{AutoML for Imbalanced learning}\label{AutoML}
AutoML \citep{Hutter2019AutomatedML} is the field of automated model selection and hyperparameter configuration of machine learning or deep learning models. The latter is commonly referred to as Neural Architecture Search (NAS) \citep{Elsken2019NeuralAS}. %AutoML reduces human effort with respect to manual selection of hyperparameters and models for the selected dataset. A simple AutoML pipeline consists of a search algorithm that selects the best model for the given dataset. We can add meta-learning to provide a search algorithm with better priors of models and hyperparameters. Few popular AutoML methods include AutoSklearn \citep{Feurer2019AutosklearnEA}, TPOT \citep{Olson2016TPOTAT}, AutoWEKA  \citep{Kotthoff2017AutoWEKA2A}, GAMA \citep{gama} etc.
The main motivation for this paper is that most AutoML systems show weak performance on imbalanced datasets. As shown in \cite{Gijsbers2019AnOS}, all evaluateed AutoML frameworks performed worse than a Random Forest on two imbalanced datasets. 
AutoML for imbalanced learning problems has caught some interest recently: \cite{MONIZ2021115011} proposed the ATOMIC framework which used meta-learning to select balancing techniques, while \cite{TPE} used Tree parzon estimators \citep{tpes} for CASH optimization on imbalanced learning problems. 
%The ATOMIC framework utilized meta learning based strategy for their workflow optimization. 
ATOMIC \citep{MONIZ2021115011} uses only one classifier (Random Forest) with multiple sampling strategies to find the optimal configuration for the imbalanced problem. \cite{TPE} uses tree Parzen estimators for both hyperparameter optimization and the selection of balancing techniques, and shows that this outperforms Random search and evolutionary algorithms.

 In this paper, we introduce the AutoBalance framework, which has a much wider search space than previously proposed approaches and can thus more precisely select the best approach. It can also use more efficient search algorithms, including ASHA \citep{asha} and AsyncEA \cite{asyncea}, together with appropriate objective functions, and uses a multi-phase approach that also includes meta-learning.

\section{Problem Definition}\label{problem}
Searching the optimal configuration of a machine learning pipeline is one of the main goal of an AutoML system. The problem is described for a fixed dataset $\boldsymbol{D} = \big\{(x^i, y^i)$, $i=1,..,n\big\}$. \textit{Combined algorithm selection and hyperparameter optimization (CASH)} \citep{Thornton2013AutoWEKACS} is the search over learning algorithms $\boldsymbol{A}$ and associated hyperparameter spaces $\boldsymbol{\Lambda}$ for an optimal combination $A^*_\lambda$ that maximizes the performance of prediction over $k$ subsets of $\boldsymbol{D}$ (e.g., $k$ cross-validation folds). Equation \ref{eq:1} formalizes this optimization problem, where $L$ is an evaluation measure, and $\big\{\boldsymbol{X}_{tr}, \boldsymbol{y}_{tr}\big\}$ and $\big\{\boldsymbol{X}_{val}, \boldsymbol{y}_{val}\big\}$ represent the training and validation sets, respectively.  The search can be extended to include preprocessing algorithms as well as postprocessing steps, in which case $\boldsymbol{A}$ is the space of all possible pipelines.

\begin{equation}
\begin{split}
A^*_{\lambda} =
\operatorname*{argmin}_{%
       \substack{%
         \forall A^j \in \boldsymbol{A} \\
         \forall \lambda \in \boldsymbol{\Lambda}
       }
     }
\frac{1}{k} 
\sum_{i=1}^{k} L \left( A^j_\lambda, \big\{\boldsymbol{X}^i_{tr}, \boldsymbol{y}^i_{tr}\big\}, \big\{\boldsymbol{X}^i_{val}, \boldsymbol{y}^i_{val}\big\} \right)
\end{split}
\label{eq:1}
\end{equation}

\par
AutoML for imbalanced learning setting can is similar to equation \ref{eq:1}. The difference between imbalanced learning setting is the ratio of majority class(es) $y_M$ to minority class(es) $y_m$. In imbalanced learning setting the number of majority class instances vs minority class instances is very high: $ \lvert y_M\rvert >>\lvert y_m\rvert$.

\section{\textit{AutoBalance}: Automated learning for imbalanced datasets}\label{AutoBalance}
\begin{figure}[t]
    \centering
    \includegraphics[scale=0.4]{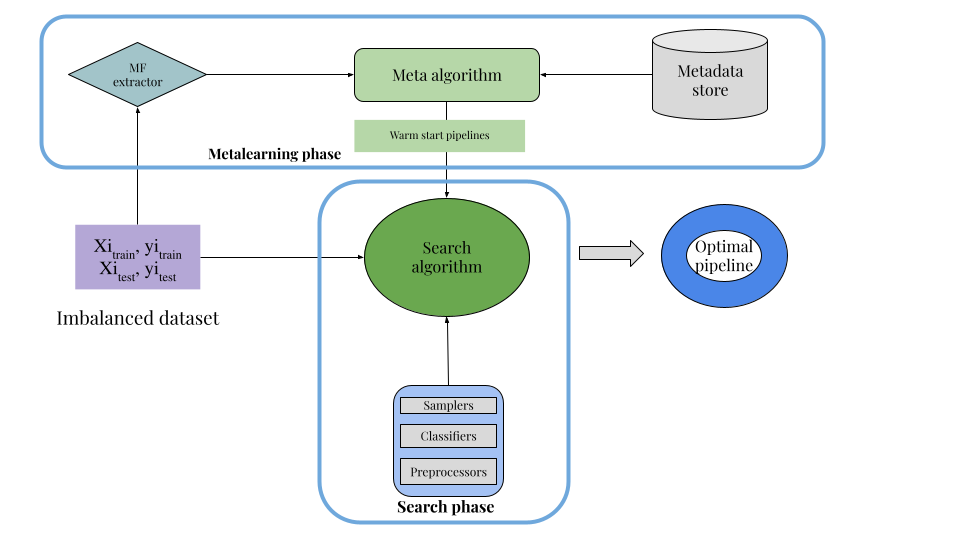}
    \caption{AutoBalance Framework}
    \label{fig:autobalance}
\end{figure}

In this paper, we propose Automated Imbalanced Learning(\textit{AutoBalance})\footnote{\url{https://github.com/prabhant/gama/tree/imblearn}}. Figure~\ref{fig:autobalance} shows an overview of the structure of our method, with the different system modules and flow. Our objective is to automate the selection and hyperparameter optimization of pipelines for Imbalanced learning tasks. Our motivation to develop this framework is to automate the selection of pipeline components from the wide array of options available for Imbalanced learning tasks. The intuition is to make it easy for machine learning practitioners to use these specialized components without having domain expertise in imbalanced machine learning. We use meta-learning for the warm start of our search algorithm. Our search algorithm finds the best  pipeline for the selected task. We implement build AutoBalance  on top of General Automated Machine learning Assistant(GAMA) \cite{gama} and use estimators from scikit-learn \citep{scikit-learn} and Imbalanced-learn \citep{Lematre2017ImbalancedlearnAP}. \par
    \textit{AutoBalance} is a full pipeline optimization system; pipelines can include one or more learning algorithms, as well as multiple preprocessing steps. Users can also add any scikit-learn compatible sampling techniques in our search space to make it wider. \textit{AutoBalance} allows for a time constraint. Another problem with Imbalance learning is the right selection of metrics. Accuracy is not the right metric for most imbalanced learning tasks. To solve this problem we integrate multiple metrics suitable for Imbalance learning like balanced accuracy, Geometric mean, F1 score, and sensitivity score. Users can optimize for any metric they wish using AutoBalance. Users can also define new metrics in \textit{AutoBalance} as long as they are compatible with scikit-learn API. We build \textit{AutoBalance}  on top of GAMA \citep{gama} library. \textit{AutoBalance} allows ensemble postprocessing as well. 
We now describe components of our system :
\subsection{Search space}\label{searchspac}
AutoBalance search space consist wide array of sampling techniques, as well as ensemble based classifiers from imbalanced learn. Our search space consist of undersampling techniques AllKNN, Cluster centroids, Edited nearest neighbours, Condensed nearest neighbours, Oversampling techniques: ADASYN, BorderlineSMOTE, SVMSMOTE, SMOTE and ensemble classifiers. To make AutoBalance more robust we include various classifiers, preprocessing techniques and feature selection techniques from scikit-learn. Details of the entire search space with hyperparameter configurations can be found on github repository\footnote{\url{https://github.com/prabhant/gama/blob/imblearn/gama/configuration/classification.py}}. To the best of our knowledge we have the largest search space among all available AutoML tools for imbalance learning. We describe search space of AutoBalance in Table \ref{tab:searchspace} (We have only included imbalance learning based components in the table because of space constraints.)

\begin{sidewaystable}[!]
\caption{Search space of AutoBalance, with the imbalance learners on top, and the main sampling methods below.}
\label{tab:searchspace}
\resizebox{0.9\linewidth}{!}{%
\begin{tabu}{llll} 
\toprule
\textbf{Model} & \textbf{Hyperparameter} & \textbf{Default value} & \textbf{Search range} \\ 
\hline
\multirow{4}{*}{Balanced Random Forest Classifier} & n\_estimators & 100 & 100 \\
 & criterion & info\_gini & \{gini, entropy\} \\
 & max\_features & Auto & {[}0.05-1.01] \\
 & min\_impurity\_decrease & 0.0 & {[}0.05-1.01] \\
\hline
\multirow{3}{*}{Balanced Bagging Classifier} & n\_estimators & 10 & {[}100] \\
 & max\_features & 1.0 & {[}0.05-1.01] \\
 & max\_samples & 1.0 & {[}0.05-1.01] \\

\hline
\multirow{2}{*}{RUSBoost Classifier} & learning\_rate & 1.0 & {[}0.05-1.01] \\
 & n\_estimators & 10 & {[}50,100] \\
\hline
\multirow{3}{*}{Borderline SMOTE} & k\_neighbours & 5 & {[}1-25] \\
 & kind & borderline-1 & {[}Borderline-1, Borderline-2] \\ 
 & m\_neighbours & 10 & {[}1-25] \\
\hline
SMOTE & k\_neighbours & 5 & {[}1-25] \\
\hline
ADASYN & k\_neighbours & 5 & {[}1-25] \\
\hline

Edited Nearest neighbour & k\_neighbours & 5 & {[}1-25] \\
\hline
Condensed Nearest Neighbour & k\_neighbours & 5 & {[}1-25] \\
\hline
AllKNN & k\_neighbours & 5 & {[}1-25] \\
\hline
Cluster Centroids & voting & auto & {[}auto, hard, soft] \\
\hline
SMOTEENN & Sampling\_strategy & auto & {[}auto, minority, all] \\
\hline
SMOTETomek & sampling\_strategy & auto & {[}1-25] \\

\hline

\bottomrule
\end{tabu}
}
\end{sidewaystable}
% \begin{sidewaystable}
% \begin{minipage}
% \begin{tabular}{c|c|c|c}
% \toprule
% \textbf{Model} & \textbf{Hyperparameter} & \textbf{Default value} & \textbf{Search range} \\ 
% \multirow{4}{*}{Balanced Random Forest Classifier} & criterion & gini & {gini, entropy}
% \hline

% \midrule
% \end{tabular}
% \end{minipage}
% \end{sidewaystable}

\subsection{Warm start phase}\label{warmstart}
We use meta-learning\citep{Vanschoren2018MetaLearningAS} to warm start \textit{AutoBalance}. The warm start approach is similar to learning to rank appraoch described in \citep{Vanschoren2018MetaLearningAS}. We use warm start optimization for similar tasks. The meta-learning  phase includes a meta-feature extractor and meta-learning algorithm. Our meta-learning algorithm uses cosine similarity between meta-features to give us the most similar datasets to the current dataset. We would like to emphasize that because of modular structure of \textit{AutoBalance} and GAMA \citep{gama}, cosine similarity can be replaced with other similarity metric or distance metric if required by user. We take the pipelines that perform the best in that data set and use them to initialize the population in \textit{AutoBalance}. An AutoML model is trained on a collection of meta datasets. Information about meta dataset, metafeatures, performance of the best found pipeline is stored in a metadata store. 
\par
We describe our warm start approach in Algorithm\ref{alg:one}. We first describe the metadataset collection as $D_m$. We query the datasets and compute Metafeatures $F_d$ via $MFE$ and best found pipelines $A^*$ for each dataset $d$ and append them to the metadata store $MS$. When user inputs a new dataset $D_n$ to AutoBalance, AutoBalance first computes the metafeatures $F_n$ and then find similarities $S_n$ with the available metadatasets from the Metadata store via similarity measure $f_s()$ . Then we select the top similarities $S_t$ from the similarity array and choose the top performing pipielines $A^*_s$ for warm starting AutoBalance.
\begin{algorithm}
\begin{algorithmic}
\caption{Pseudocode for warm start phase}\label{alg:one}
\Require \State $MFE \gets Meta Feature Extractor$
\State $D_m \gets MetaDatasets$ 
\For{\texttt{d in $D_m$}}
    \State $F_d \gets MFE(d)$ \Comment{Computing metafeatures via metafeature extractor}
    \State $A^* \gets AutoBalance(d)$
    \State $MS \gets [F_d, A^*]$ \Comment{Populating Metadata Store}
\EndFor
\State $D_n \gets New Dataset$
\State $F_n \gets MFE(D_n)$
\For{\texttt{i in $MS$}}
\State $Sn \gets f_s(F_n, F_i)$
\EndFor
\State $S_t \gets S[Top_m]$
\State $A^*_S \gets MS(S_t, A*)$
\State $A^*_n \gets AutoBalance(A^*_S)$ \Comment{Best performing pipeline}
\end{algorithmic}
\end{algorithm}

\subsection{Search phase}\label{subsec2}
We use GAMA's genetic programming configuration and search algorithms as one of our optimization methods.  $AutoBalance{Search}$, can be \textit{Asynchronous Evolutionary Optimization} \citep{asyncea}, \textit{Random Search} \citep{JMLR:v13:bergstra12a} or \textit{Asynchronous Successive Halving (ASHA)} \citep{asha}.

\section{Benchmarks}\label{benchmarks}

A major problem with Imbalanced learning is the lack of collection of datasets with imbalanced benchmarks. Existing benchmarks consist of Zenodo benchmark with very limited number of datasets, KEEL \citep{FERNANDEZ20082378} dataset repository does provide with a number of Imbalanced datasets with different ratio but suffers heavily from data duplicates. Other works in Imbalanced learning use their own collection of selected datasets from open source domain. \cite{survey} also mentions lack of repositories with imbalanced datasets. To solve this inconsistency of Imbalanced datasets, we propose four dataset benchmarks in this work:
\begin{enumerate}

\item Imbalanced Binary Classification Benchmark.(32 datasets, Table \ref{tab:binarybenchmark})
\item Extremely Imbalanced Binary Classification Benchmark.(11 datasets, Table \ref{tab:extremebinarybenchmark})
\item Imbalanced Multiclass Classification Benchmark.(28 datasets, Table \ref{tab:multiclassbenchmark})
\item Extremely Imbalanced Multiclass Classification Benchmark.(22 datasets, Table \ref{tab:exmulticlassbenchmark})

\end{enumerate}
\par
We use OpenML \citep{OpenML2013} to get these datasets. 
We assume that Extreme imbalance of classes and imbalance of classes are two different problems that should be solved separately. We define extreme Imbalance as the ratio of majority to minority class of at least 20:1 and imbalanced datasets as the majority to minority class ratio of at least 3:1. To avoid duplicate of results we do not include extremely imbalanced datasets in the Imbalanced benchmarks. Datasets included are also required to have at least two samples of minority class. We use the OpenML benchmark suites \citep{Bischl2021OpenMLBS}  framework to design the datasets.
Our criteria for quality control on these benchmarks is:

\begin{enumerate}
    \item No duplicates.
    \item No alternate version of datasets with different class distribution.
    \item Datasets must be verified on OpenML
    \item There should be at least two samples of the minority class.
\end{enumerate}
\par
 The list of all the datasets in the benchmarks can be found in the Appendix A \ref{appendixa}.

\section{Experiments and Results}\label{experiments}
In this section we desribe our experimental setup for evaluating AutoBalance and report and analyse the results. 

\subsection{Experimental setup}
For our experiments we run AutoBalance with selected search space.  The features extracted by Metafeature Extractor \citep{Alcobaa2020MFETR} can be found in Appendix B \ref{appendixb}.
%% metafeatures list here

We run AutoBalance for one hour. For this experiment we use ten warm start candidates, the number of warm start candidates is chosen intuitively now but in future studies can be conducted to decide the right number of warm start candidates. We run these experiments on one core to ensure consistency. We conduct 4 sets of experiments, one on every dataset benchmark we presented. 
For warm start, we train pipelines on OpenMLCC18 benchmark \footnote{\url{https://www.openml.org/s/99}} with AutoBalance(without warmstarting) and save best performing pipelines with metafeatures in our metadata store. We make sure that there is no overlap between meta datasets and datasets on which AutoBalance will be evaluated(There were 6 datasets which were present in both OpenML cc18 and our propsed benchmarks). We use cosine similarity as a similarity metric between metafeatures. AutoBalance is initialised with 10 warm start candidates. For current set of experiments we use AsyncEA as our search algorithm. 
\par
We compare our results with AutoSklearn \citep{Feurer2019AutosklearnEA} with a one hour time budget for searching pipelines. We use balanced accuracy as our evaluation metric.
There are different sources that promote different definitions of balanced accuracy. We use the implementation of balanced accuracy from scikit-learn \citep {scikit-learn, Kelleher2015FundamentalsOM, balanced_acc}
\par

We report the balanced accuracy on dataset from Imbalanced Binary Classification Benchmark in Table \ref{tab:binary}, from Extremely Imbalanced Binary Classification Benchmark in Table \ref{tab:exbinary}, Imbalanced Multiclass Classification Benchmark in Table \ref{tab:multi} and Extremely Imbalanced Multiclass Classification Benchmark in Table \ref{tab:exmulti}

\begin{table}[]
    \centering
\begin{tabular}{c|c|c}
\toprule
           Dataset &  AutoBalance &    Auto-Sklearn \\
           \midrule
           PizzaCutter1 &           \textbf{0.727753} &  0.570387 \\
               Stagger1 &           1.000000 &  1.000000 \\
                    kc3 &           \textbf{0.781469} &  0.545455 \\
             confidence &           0.600000 &  0.833333 \\
            sylva\_prior &          \textbf0.993930{}  &  0.984562 \\
                    pc4 &           \textbf{0.855035} &  0.692361 \\
      ipums\_la\_99-small &           \textbf{0.817722} &  0.526236 \\
         mfeat-karhunen &           \textbf{0.998889} &  0.990000 \\
                    jm1 &           \textbf{0.678592} &  0.564777 \\
                    pc3 &           \textbf{0.797115} &  0.536075 \\
            page-blocks &           0.947018 &  0.944711 \\
            mfeat-pixel &           0.990000 &  0.996667 \\
                    ar6 &           0.352273 &  0.500000 \\
      synthetic\_control &           1.000000 &  1.000000 \\
                    ar4 &           0.677273 &  0.854545 \\
          mfeat-zernike &           0.997778 &  0.997778 \\
            hypothyroid &           0.991427 &  0.993151 \\
         JapaneseVowels &           0.978240 &  0.988422 \\
          mfeat-factors &           \textbf{1.000000} &  0.987778 \\
      ipums\_la\_98-small &           \textbf{0.807348} &  0.544993 \\
   analcatdata\_birthday &           \textbf{0.879260} &  0.769718 \\
 analcatdata\_halloffame &           \textbf{0.911503} &  0.799873 \\
                segment &           0.987952 &  0.998990 \\
           spectrometer &           \textbf{0.970588} &  0.911765 \\
 arsenic-female-bladder &           \textbf{0.783333} &  0.633333 \\
                   sick &           \textbf{0.968634} &  0.927080 \\
    analcatdata\_lawsuit &           1.000000 &  1.000000 \\
  visualizing\_livestock &           \textbf{0.740741} &  0.481481 \\
     analcatdata\_apnea2 &           0.908374 &  0.906250 \\
     analcatdata\_apnea3 &           \textbf{0.928932} &  0.908369 \\
                   meta &           \textbf{0.791162} &  0.491597 \\
                  scene &           \textbf{0.973815} &  0.967593 \\
\bottomrule
\end{tabular}
    \caption{Balanced accuracy of AutoBalance vs AutoSklearn for Imbalanced Binary Classification Benchmark, bold balanced accuracy indicate win.}
    \label{tab:binary}
\end{table}

\begin{table}[]
    \centering
\begin{tabular}{c|c|c}
\toprule
         Dataset &  AutoBalance &    Auto-Sklearn \\
         \midrule
 arsenic-male-bladder &           0.671642 &  0.833333 \\
            PieChart2 &           \textbf{0.833333} &  0.497268 \\
            oil\_spill &           \textbf{0.714444} &  0.595556 \\
                  mc1 &           \textbf{0.803154} &  0.646633 \\
               Speech &           \textbf{0.633039} &  0.533333 \\
                  pc2 &           \textbf{0.880388} &  0.500000 \\
           APSFailure &           \textbf{0.963059} &  0.889026 \\
                  dis &           \textbf{0.962823} &  0.599461 \\
           creditcard &           \textbf{0.926413} &  0.865741 \\
          mammography &           \textbf{0.893967} &  0.805861 \\
            yeast\_ml8 &           \textbf{0.869780} &  0.500000 \\
\bottomrule
\end{tabular}
    \caption{Balanced accuracy of AutoBalance vs AutoSklearn for Extremely Imbalanced Binary Classification Benchmark, bold balanced accuracy indicate win.}
    \label{tab:exbinary}
\end{table}

\begin{table}[]
    \centering
\begin{tabular}{c|c|c}
\toprule
                               \textbf{Dataset name} &  AutoBalance &    Auto-sklearn \\
                               \midrule

                                  connect-4 &           \textbf{0.713319} &  0.675420 \\
                                      flags &           0.356448 &  0.354563 \\
     jungle\_chess\_2pcs\_endgame\_panther\_lion &           1.000000 &  1.000000 \\
                               prnn\_viruses &           1.000000 &  1.000000 \\
                   analcatdata\_broadwaymult &           \textbf{0.346032} &  0.300000 \\
                           autoUniv-au7-500 &           \textbf{0.335684} &  0.302309 \\
                                thyroid-new &           0.867725 &  0.887218 \\
              meta\_instanceincremental.arff &           \textbf{}0.589286 &  0.482143 \\
 jungle\_chess\_2pcs\_endgame\_panther\_elephant &           0.998675 &  0.999337 \\
                 meta\_batchincremental.arff &           \textbf{0.322115} &  0.250000 \\
                                      ecoli &           \textbf{0.798709} &  0.765488 \\
                            squash-unstored &           0.500000 &  0.500000 \\
                                    bridges &           0.494949 &  0.636364 \\
                          microaggregation2 &           \textbf{0.500240} &  0.374659 \\
                                    nursery &           0.998126 &  1.000000 \\
                          ipums\_la\_97-small &           \textbf{0.361693} &  0.325386 \\
         jungle\_chess\_2pcs\_endgame\_rat\_lion &           \textbf{0.990779} &  0.981793 \\
                                    volkert &           \textbf{0.625588} &  0.538242 \\
         jungle\_chess\_2pcs\_endgame\_complete &           0.991661 &  0.994870 \\
                         robot-failures-lp3 &           0.387500 &  0.437500 \\
                      wall-robot-navigation &           0.998592 &  0.999523 \\
                                    collins &           \textbf{0.247831} &  0.217378 \\
                                prnn\_fglass &           \textbf{0.500731} &  0.460404 \\
                                      glass &           \textbf{0.723806} &  0.681043 \\
                                 micro-mass &           0.888475 &  0.933654 \\
                           cardiotocography &           1.000000 &  1.000000 \\
                                    heart-h &           0.193465 &  0.194529 \\
                          autoUniv-au4-2500 &           0.461897 &  0.472445 \\
\bottomrule
\end{tabular}
    \caption{Balanced accuracy of AutoBalance vs AutoSklearn for Imbalanced Multiclass Classification Benchmark, bold balanced accuracy indicate win.}
    \label{tab:multi}
\end{table}

\begin{table}[htb]
    \centering
\begin{tabular}{c|c|c}
\toprule
               Dataset name &  AutoBalance &    Auto-sklearn \\
\midrule

                thyroid-dis &           \textbf{0.579901} &  0.426047 \\
                    shuttle &           \textbf{0.952368} &  0.856955 \\
           walking-activity &           0.607113 &  0.628591 \\
                thyroid-ann &           \textbf{0.996942} &  0.984360 \\
                     anneal &           0.976608 &  1.000000 \\
                      lymph &           0.679167 &  0.895833 \\
                     helena &           \textbf{0.238234} &  0.214591 \\
                     jannis &           \textbf{0.607618} &  0.565518 \\
               Indian\_pines &           0.873247 &  0.936270 \\
                   baseball &           \textbf{0.722551} &  0.560648 \\
                      yeast &           \textbf{0.507521} &  0.401817 \\
                page-blocks &           \textbf{0.914376} &  0.840056 \\
                      allbp &           \textbf{0.861094} &  0.630666 \\
                  covertype &           \textbf{0.873699} &  0.243437 \\
     analcatdata\_halloffame &           \textbf{0.734668} &  0.560648 \\
                       ldpa &           \textbf{0.870157} &  0.697121 \\
           wine-quality-red &           \textbf{0.492832} &  0.310061 \\
 meta\_stream\_intervals.arff &           0.954184 &  0.972484 \\
           thyroid-allhyper &           \textbf{0.596032} &  0.426047 \\
                    kr-vs-k &           0.766596 &  0.857321 \\
                     allrep &           \textbf{0.840278} &  0.700581 \\
                      kropt &           0.749832 &  0.900749 \\
\bottomrule
\end{tabular}
    \caption{Balanced accuracy of AutoBalance vs AutoSklearn for Extremely Imbalanced Multiclass Classification Benchmark, bold balanced accuracy indicate win.}
    \label{tab:exmulti}
\end{table}

\subsection{Results and Discussion}\label{analysis}
In this section we discuss the results of our experiments. We first analyse the performance of AutoBalance, then we discuss the pipelines AutoBalance discovered during the experiments. 

\subsubsection{Performance Analysis}

In this section we describe our observations and results of our experiments and performance of AutoBalance. For analysis in this work we describe a win if balanced accuracy of AutoBalance is greater than AutoSklearn by more than one percent. Draw happens when AutoBalance performance is equal to or in the range of one percent with Autosklearn. AutoBalance loses if the balanced accuracy of AutoSklearn is greater than AutoBalance by more than one percent.

\begin{enumerate}
    \item For Imbalanced Binary Classification Benchmark, AutoBalance won 16 times, autosklearn and AutoBalance draw were 11 times and AutoBalance lost 5 times.
    \item For the Extremely Imbalanced Binary Classification Benchmark, benchmark AutoBalance wins 10 times and loses one time.
    \item For the Imbalanced Multiclass Classification Benchmark AutoBalance wins 12 times, draws 12 times, and loses 4 times.
    \item For Extremely Imbalanced Multiclass Classification Benchmark: AutoBalance won 15 times and lost 7 times.
\end{enumerate}
     
\begin{figure}[htb]
    \centering
    \includegraphics[scale=0.4]{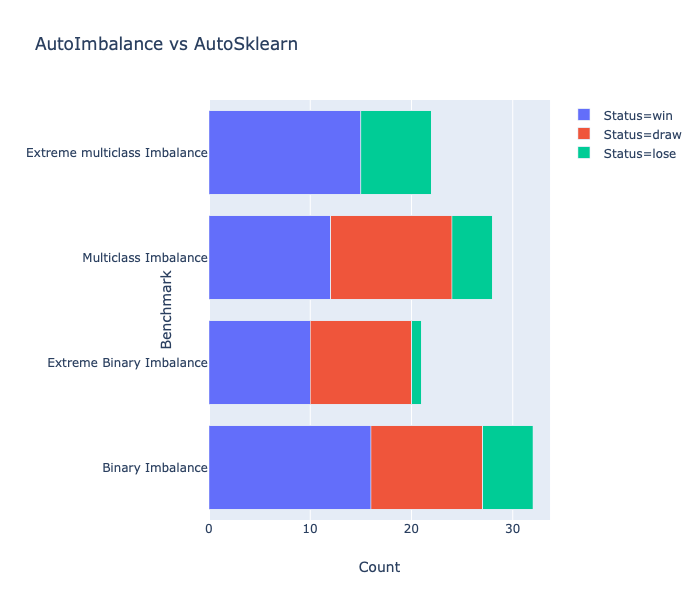}
    \caption{AutoBalance vs AutoSklearn performance}
    \label{fig:barchart}
\end{figure}

These results are described in a more visual manner in Figure \ref{fig:barchart}.
We observed that a number of draws occurred when the performance of classifiers was really high(0.98-1.00 balanced accuracy). 

AutoBalance outperforms Auto-sklearn on a number of tasks. We observed that most of the draws happen when the dataset was too easy and the balanced accuracy achieved was really high. In imbalanced multiclass and extremely imbalanced multiclass benchmarks AutoBalance performance was relatively poor than AutoBalance performance on imbalanced binary and extremely imbalanced binary benchmarks(It still outperformed Auto-Sklearn). This can be due to a number of factors like time limit, pipeline evaluation time limit(our default pipeline evaluation time is max(0.1 x time budget)), search algorithms and lack of multiclass imbalanced sampling techniques like SOUP \citep{Grycza2020multiimbalanceOS}. This limitation can be resolved by integrating samplers from libraries like \cite{Grycza2020multiimbalanceOS, Koziarski2021CSMOUTECS}. We attempted to use these libraries with AutoBalance but due to lack of scikit-learn like compatibility we were not able to use them. In future they can use integrated if we modify these libraries to be compatible with the API. This can allow us to have an improved version of AutoBalance for multiclass problems. 

\subsubsection{Pipeline Properties}
We wanted to know more about the selection criteria and design of search space for future research questions. This section includes the most common classifiers and preprocessors that were selected as the result of classifiers. \par
We include pipeline components in Table \ref{tab3} . We observe that Balanced bagging classifier \citep{Breiman2004BaggingP} is the most common estimator among selected optimal pipelines. Samplers are scattered all over the selected pipelines pool; this observation can emphasize the importance of using AutoBalance as different datasets require different sampling techniques. We observed that estimators did require some sort of undersampling technique. 
\par
\begin{figure}[htb]
    \centering
    \includegraphics[scale=0.4]{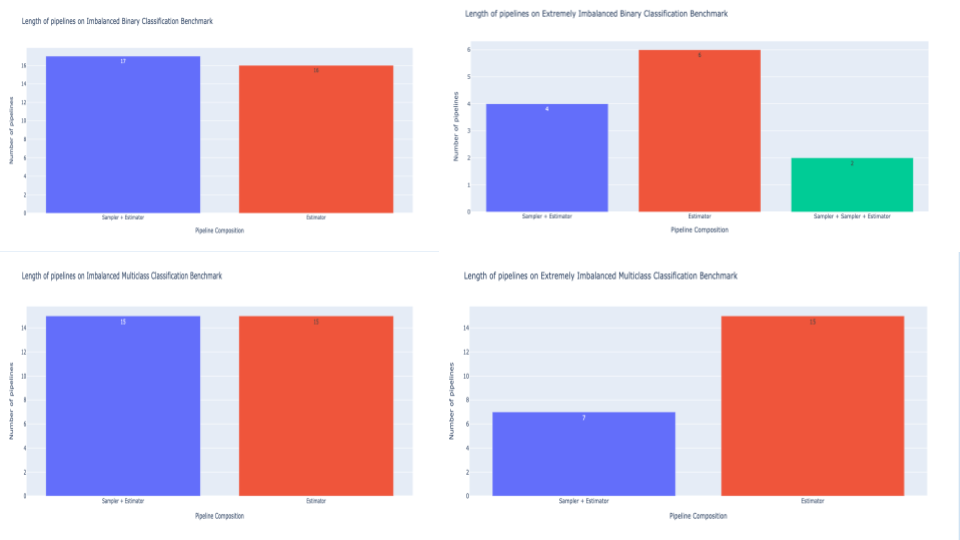}
    \caption{Pipeline Composition summary on different benchmarks}
    \label{fig:pipeline}
\end{figure}

 We summarize the pipeline composition of different sampling and preprocessing techniques with the estimator in Figure \ref{fig:pipeline}. We would like to reiterate that in the current experiment AutoBalance had a limited time budget, and the maximum amount of time for evaluating per pipeline was limited to ten percent of the time budget. If user increases the time budget or pipeline evaluation time then the length of pipelines can vary. 

\bmhead{Ablation Studies} To determine how individual parts influenced the performance of our proposed method we conducted an ablation study. Our ablation study checks the influence of search space over in AutoBalance. For this experiment we run AutoBalance without warm starting for 3600 seconds. We run AutoBalance on Extremely Imbalanced Binary Classification Benchmark with Random Search, AsychEA and Asynchronous Succesive Halving. We compare the results with AutoBalance without Imbalanced-learn estimators and samplers(The search space is similar to GAMA search space then with minor changes). 
We report the results of this study in Table \ref{tab:ablation} . These results indicate that having Imbalanced learning samplers and estimators is an advantage over a scikit-learn only search space. 
\begin{table}[]

    \centering

    \begin{tabular}{c|c|c|c}
    \toprule
         Search Algorithm & Win & Draw & Lose \\
         \midrule
         Random Search & \textbf{6} & 0 & 5 \\
         ASHA & \textbf{7} & 0 & 4 \\
         AsyncEA & \textbf{5} & 3 & 3 \\
    \end{tabular}
    \caption{Peroformance of AutoBalance with and without Imbalanced learning based classifiers on Extremely Imbalanced Binary Benchmark}
    
    \label{tab:ablation}
\end{table}
% Binary imbalance
\begin{sidewaystable}
\sidewaystablefn%
\begin{center}
\begin{minipage}{\textheight}
\caption{Occurences of different samplers and classifiers in the discovered pipelines separated by benchmark type}\label{tab3}
\begin{tabular*}{\textheight}{@{\extracolsep{\fill}}lcccc@{\extracolsep{\fill}}}
{Imbalanced Benchmarks} &  \textbf{Binary } & \textbf{E.Binary }  &  \textbf{Multiclass } & \textbf{E.Multiclass }  \\
\textbf{Components}                        &                   &                           &                       &                               \\
\midrule
\textbf{ADASYN()}                         &               2 &                       0 &                   0 &                           0 \\
\textbf{BalancedBaggingClassifier()}      &              14 &                       8 &                  19 &                          16 \\
\textbf{PCA()}                            &               2 &                       0 &                   2 &                           1 \\
\textbf{EasyEnsembleClassifier()}         &               3 &                       0 &                   0 &                           1 \\
\textbf{RandomForestClassifier()}         &               4 &                       0 &                   6 &                           4 \\
\textbf{EditedNearestNeighbours()}        &               1 &                       1 &                   0 &                           0 \\
\textbf{RUSBoostClassifier()}             &               6 &                       1 &                   0 &                           1 \\
\textbf{FastICA()}                        &               2 &                       0 &                   1 &                           1 \\
\textbf{KNeighborsClassifier() }          &               3 &                       1 &                   0 &                           0 \\
\textbf{SMOTEENN() }                      &               1 &                       0 &                   0 &                           0 \\
\textbf{ClusterCentroids()}               &               1 &                       0 &                   0 &                           0 \\
\textbf{LogisticRegression()}             &               2 &                       0 &                   1 &                           0 \\
\textbf{BorderlineSMOTE()}                &               1 &                       0 &                   3 &                           2 \\
\textbf{PolynomialFeatures()}             &               1 &                       1 &                   1 &                           0 \\
\textbf{AllKNN() }                        &               1 &                       1 &                   0 &                           0 \\
\textbf{Nystroem() }                      &               1 &                       0 &                   0 &                           0 \\
\textbf{Normalizer()}                     &               1 &                       1 &                   1 &                           0 \\
\textbf{BalancedRandomForestClassifier()} &               1 &                       2 &                   0 &                           0 \\
\textbf{CondensedNearestNeighbour()}      &               2 &                       1 &                   0 &                           0 \\
\textbf{SMOTE()}                          &               1 &                       0 &                   5 &                            \\
\textbf{Binarizer()}                      &               0 &                       1 &                   2 &                           0 \\
\textbf{VarianceThreshold()}              &               0 &                       1 &                   0 &                           0 \\
\textbf{SVMSMOTE()}                       &               0 &                       1 &                   0 &                           1 \\
\textbf{GradientBoostingClassifier()}     &               0 &                       0 &                   4 &                           0 \\
\bottomrule
\end{tabular*}
\footnotetext{Note: E.Binary refers to Extreme binary and E.Multiclass refers to Extreme Multiclass Imbalanced benchmarks.  These components can change with search algorithm choice and objective provided. Some larger pipelines were ignored by AutoBalance due to time budget.}
\end{minipage}
\end{center}
\end{sidewaystable}
\section{Conclusion and Future Work}\label{conclusion}
In this work we presented a novel AutoML framework for Imbalanced learning tasks called \textbf{AutoBalance} which consist of a meta learning phase and a search phase. AutoBalance can take advantage of different metalearning methods and search algorithms for searching pipeline for imbalanced learning tasks. AutoBalance is a full pipeline optimization tool that leverages data preprocessing techniques and ensemble methods available for imbalanced tasks. As a result, it allows end users of the AutoML tool to utilize these specialized components without the expertise in imbalanced learning. 

\par
We proposed four benchmarks for imbalanced learning tasks. We aim to ensure that these benchmarks can serve as a standard for future research on imbalanced learning research and AutoML research.
\par
We presented analysis of our experiments in which AutoBalance outperformed state of the art AutoML methods on a variety of tasks. We presented most common winner pipeline components as well.
\par
AutoBalance is currently one of the few AutoML frameworks in the area of Imbalance learning. Future work includes incorporating our benchmark with the AutoML benchmark \citep{Gijsbers2019AnOS} for Imbalance learning and compare more AutoML frameworks with each other. Another extension of AutoBalance would be to support imbalanced regression problems \citep{Ribeiro2020ImbalancedRA} which is a much less studied domain. We would also like to incorporate more advanced meta-learning methods in AutoBalance, such as Optimal transport and Collaborative filtering. Another interesting research direction would be to integrate AutoBalance with an online AutoML framework like OAML \citep{Celik2022OnlineAA} to extend this research to imbalanced data streams. Finally, we aim to use AutoBalance and our proposed benchmarks as a large scale study to evaluate the best combinations of pipelines and samplers, such as integrating AutoBalance with SMOTE-variants \citep{smote-variants} to evaluate the best combination of smote variants with any estimator. 
\section*{Declarations}

\begin{itemize}
\item Funding: This research was supported by the European Commission's H2020 program under the StairwAI grant.

\item Availability of data and materials and code: All our data, materials and code are available publicly at \url{https://github.com/prabhant/gama/tree/imblearn}. All the datasets are avaiable on OpenML.

\item Financial or non-financial interests: Not applicable.
\item Conflicts of interest/Competing interests: No conflicts
\item Ethics approval: Not applicable.

\item Consent to participate: Not applicable.
\item Consent for publication - Not applicable
\item Authors' contributions: Prabhant Singh contributed to the conceptualisation, development, experiments and writing of this work, Joaquin Vanschoren contributed to conceptualisation and writing. 
\end{itemize}

\newpage
\bmhead{Acknowledgments}
We would like to give special thanks to Pieter Gijsbers and Bilge Celik for their suggestions while writing this paper. This research was supported by the European Commission's H2020 program under the StairwAI grant.
\bibliography{references}% common bib file
\newpage
\section{Appendix A}\label{appendixa}
Imbalanced benchmarks from OpenML
\begin{sidewaystable}
\sidewaystablefn%
\begin{tabular}{lrrrr}

\toprule
                   \textbf{Name} &  \textbf{MajorityClassSize} &  \textbf{MinorityClassSize} &  \textbf{NumberOfFeatures} &  \textbf{NumberOfInstances} \\
\midrule
                   sick &             3541.0 &              231.0 &              30.0 &             3772.0 \\
                  scene &             1976.0 &              431.0 &             300.0 &             2407.0 \\
    analcatdata\_lawsuit &              245.0 &               19.0 &               5.0 &              264.0 \\
                   meta &              474.0 &               54.0 &              22.0 &              528.0 \\
     analcatdata\_apnea3 &              395.0 &               55.0 &               4.0 &              450.0 \\
     analcatdata\_apnea2 &              411.0 &               64.0 &               4.0 &              475.0 \\
  visualizing\_livestock &              105.0 &               25.0 &               3.0 &              130.0 \\
 arsenic-female-bladder &              479.0 &               80.0 &               5.0 &              559.0 \\
           spectrometer &              476.0 &               55.0 &             102.0 &              531.0 \\
                segment &             1980.0 &              330.0 &              20.0 &             2310.0 \\
 analcatdata\_halloffame &             1215.0 &              125.0 &              17.0 &             1340.0 \\
   analcatdata\_birthday &              312.0 &               53.0 &               4.0 &              365.0 \\
         JapaneseVowels &             8347.0 &             1614.0 &              15.0 &             9961.0 \\
          mfeat-factors &             1800.0 &              200.0 &             217.0 &             2000.0 \\
          mfeat-zernike &             1800.0 &              200.0 &              48.0 &             2000.0 \\
            hypothyroid &             3481.0 &              291.0 &              30.0 &             3772.0 \\
      ipums\_la\_98-small &             6694.0 &              791.0 &              56.0 &             7485.0 \\
      synthetic\_control &              500.0 &              100.0 &              61.0 &              600.0 \\
             confidence &               60.0 &               12.0 &               4.0 &               72.0 \\
      ipums\_la\_99-small &             8276.0 &              568.0 &              57.0 &             8844.0 \\
         mfeat-karhunen &             1800.0 &              200.0 &              65.0 &             2000.0 \\
            page-blocks &             4913.0 &              560.0 &              11.0 &             5473.0 \\
            mfeat-pixel &             1800.0 &              200.0 &             241.0 &             2000.0 \\
            sylva\_prior &            13509.0 &              886.0 &             109.0 &            14395.0 \\
                    pc4 &             1280.0 &              178.0 &              38.0 &             1458.0 \\
                    pc3 &             1403.0 &              160.0 &              38.0 &             1563.0 \\
                    jm1 &             8779.0 &             2106.0 &              22.0 &            10885.0 \\
                    ar4 &               87.0 &               20.0 &              30.0 &              107.0 \\
                    ar6 &               86.0 &               15.0 &              30.0 &              101.0 \\
                    kc3 &              415.0 &               43.0 &              40.0 &              458.0 \\
               Stagger1 &           888391.0 &           111609.0 &               4.0 &          1000000.0 \\
           PizzaCutter1 &              609.0 &               52.0 &              38.0 &              661.0 \\
\bottomrule
\end{tabular}
\caption{Imbalanced Binary Classification Benchmark}
\label{tab:binarybenchmark}
\end{sidewaystable}
\begin{sidewaystable}
\sidewaystablefn%
\centering
\begin{tabular}{lrrrr}
\toprule
                 \textbf{Name} &  \textbf{MajorityClassSize} &  \textbf{MinorityClassSize} &  \textbf{NumberOfFeatures} &  \textbf{NumberOfInstances} \\
\midrule
          mammography &            10923.0 &              260.0 &               7.0 &            11183.0 \\
            oil\_spill &              896.0 &               41.0 &              50.0 &              937.0 \\
            yeast\_ml8 &             2383.0 &               34.0 &             117.0 &             2417.0 \\
 arsenic-male-bladder &              535.0 &               24.0 &               5.0 &              559.0 \\
                  mc1 &             9398.0 &               68.0 &              39.0 &             9466.0 \\
                  pc2 &             5566.0 &               23.0 &              37.0 &             5589.0 \\
            PieChart2 &              729.0 &               16.0 &              37.0 &              745.0 \\
           creditcard &           284315.0 &              492.0 &              31.0 &           284807.0 \\
                  dis &             3714.0 &               58.0 &              30.0 &             3772.0 \\
               Speech &             3625.0 &               61.0 &             401.0 &             3686.0 \\
           APSFailure &            74625.0 &             1375.0 &             171.0 &            76000.0 \\
\bottomrule
\end{tabular}
    \caption{Extremely Imbalanced Binary Classification Benchmark}
    \label{tab:extremebinarybenchmark}
\end{sidewaystable}
\begin{sidewaystable}
\sidewaystablefn%
\centering
\begin{tabular}{lrrrr}
\toprule
                                       \textbf{Name} &  \textbf{MajorityClassSize} &  \textbf{MinorityClassSize} &  \textbf{NumberOfFeatures} &  \textbf{NumberOfInstances} \\
\midrule
                                      glass &               76.0 &                9.0 &              10.0 &              214.0 \\
                 meta\_batchincremental.arff &               50.0 &                3.0 &              63.0 &               74.0 \\
              meta\_instanceincremental.arff &               54.0 &                3.0 &              63.0 &               74.0 \\
                                      flags &               60.0 &                4.0 &              29.0 &              194.0 \\
                                    bridges &               44.0 &               10.0 &              12.0 &              105.0 \\
                            squash-unstored &               24.0 &                4.0 &              24.0 &               52.0 \\
                          ipums\_la\_97-small &             1938.0 &              258.0 &              61.0 &             7019.0 \\
                   analcatdata\_broadwaymult &              118.0 &               21.0 &               8.0 &              285.0 \\
                               prnn\_viruses &               39.0 &                3.0 &              19.0 &               61.0 \\
                                prnn\_fglass &               76.0 &                9.0 &              10.0 &              214.0 \\
                           cardiotocography &              579.0 &               53.0 &              36.0 &             2126.0 \\
                      wall-robot-navigation &             2205.0 &              328.0 &              25.0 &             5456.0 \\
                                 micro-mass &               60.0 &               11.0 &            1301.0 &              571.0 \\
                         robot-failures-lp3 &               20.0 &                3.0 &              91.0 &               47.0 \\
                          autoUniv-au4-2500 &             1173.0 &              196.0 &             101.0 &             2500.0 \\
                           autoUniv-au7-500 &              192.0 &               43.0 &              13.0 &              500.0 \\
                                    heart-h &              188.0 &               15.0 &              14.0 &              294.0 \\
                                    nursery &             4320.0 &              328.0 &               9.0 &            12958.0 \\
                                  connect-4 &            44473.0 &             6449.0 &              43.0 &            67557.0 \\
                                      ecoli &              143.0 &               20.0 &               8.0 &              327.0 \\
                                thyroid-new &              150.0 &               30.0 &               6.0 &              215.0 \\
                                    collins &               80.0 &                6.0 &              24.0 &             1000.0 \\
     jungle\_chess\_2pcs\_endgame\_panther\_lion &             2523.0 &              145.0 &              47.0 &             4704.0 \\
 jungle\_chess\_2pcs\_endgame\_panther\_elephant &             2495.0 &              195.0 &              47.0 &             4704.0 \\
         jungle\_chess\_2pcs\_endgame\_complete &            23062.0 &             4335.0 &              47.0 &            44819.0 \\
         jungle\_chess\_2pcs\_endgame\_rat\_lion &             3078.0 &              380.0 &              47.0 &             5880.0 \\
                                    volkert &            12806.0 &             1361.0 &             181.0 &            58310.0 \\
                          microaggregation2 &            11162.0 &              743.0 &              21.0 &            20000.0 \\
\bottomrule
\end{tabular}
    \caption{Imbalanced Multiclass Classification Benchmark}
    \label{tab:multiclassbenchmark}
\end{sidewaystable}

\begin{sidewaystable}
\sidewaystablefn%
\centering
\begin{tabular}{lrrrr}
\toprule
                       \textbf{Name} &  \textbf{MajorityClassSize} &  \textbf{MinorityClassSize} &  Numb\textbf{}erOfFeatures &  \textbf{NumberOfInstances} \\
\midrule
                     anneal &              684.0 &                8.0 &              39.0 &              898.0 \\
                      lymph &               81.0 &                2.0 &              19.0 &              148.0 \\
                page-blocks &             4913.0 &               28.0 &              11.0 &             5473.0 \\
                  covertype &           283301.0 &             2747.0 &              55.0 &           581012.0 \\
                      yeast &              463.0 &                5.0 &               9.0 &             1484.0 \\
                      kropt &             4553.0 &               27.0 &               7.0 &            28056.0 \\
                   baseball &             1215.0 &               57.0 &              17.0 &             1340.0 \\
 meta\_stream\_intervals.arff &            23021.0 &               73.0 &              75.0 &            45164.0 \\
     analcatdata\_halloffame &             1215.0 &               57.0 &              17.0 &             1340.0 \\
                    kr-vs-k &             4553.0 &               27.0 &               7.0 &            28056.0 \\
                       ldpa &            54480.0 &             1381.0 &               8.0 &           164860.0 \\
           walking-activity &            21991.0 &              911.0 &               5.0 &           149332.0 \\
           thyroid-allhyper &             1632.0 &               31.0 &              27.0 &             2800.0 \\
                thyroid-dis &             1632.0 &               31.0 &              27.0 &             2800.0 \\
                thyroid-ann &             3488.0 &               93.0 &              22.0 &             3772.0 \\
                    shuttle &            45586.0 &               10.0 &              10.0 &            58000.0 \\
           wine-quality-red &              681.0 &               10.0 &              12.0 &             1599.0 \\
                      allbp &             3609.0 &               14.0 &              30.0 &             3772.0 \\
                     allrep &             3648.0 &               34.0 &              30.0 &             3772.0 \\
                     jannis &            38522.0 &             1687.0 &              55.0 &            83733.0 \\
                     helena &             4005.0 &              111.0 &              28.0 &            65196.0 \\
               Indian\_pines &             4050.0 &               20.0 &             221.0 &             9144.0 \\
\bottomrule
\end{tabular}
    \caption{Extremely Imbalanced Multiclass Classification Benchmark}
    \label{tab:exmulticlassbenchmark}
\end{sidewaystable}
\newpage
\section{Appendix B}\label{appendixb}
Features extracted by Metafeature extractor(pyMFE) \citep{Alcobaa2020MFETR}:
\begin{enumerate}
\item AutoCorrelation
\item Cfs Subset Eval Decision Stump AUC
\item Cfs Subset Eval Decision Stump ErrRate
\item Cfs Subset Eval Decision Stump Kappa
\item Cfs Subset Eval NaiveBayes AUC
\item Cfs Subset Eval NaiveBayes ErrRate
\item Cfs Subset Eval NaiveBayes Kappa
\item Cfs Subset Eval kNN1N AUC
\item Cfs Subset Eval kNN1N ErrRate
\item Cfs Subset Eval kNN1N Kappa
\item Class Entropy
\item Decision Stump AUC
\item Decision Stump Err Rate
\item Decision Stump Kappa
\item Dimensionality
\item Equivalent Number Of Atts
\item J48.00001.AUC
\item J48.00001.ErrRate
\item J48.00001.Kappa
\item J48.0001.AUC
\item J48.0001.ErrRate
\item J48.0001.Kappa
\item J48.001.AUC
\item J48.001.ErrRate
\item J48.001.Kappa
\item Majority Class Percentage
\item Majority Class Size
\item Max Attribute Entropy
\item Max Mutual Information
\item Max Nominal Att DistinctValues
\item Mean Attribute Entropy
\item Mean Mutual Information
\item Mean Noise To SignalRatio
\item Mean Nominal Att DistinctValues
\item Min Attribute Entropy
\item Min Mutual Information
\item Min Nominal Att DistinctValues
\item Minority Class Percentage
\item Minority ClassSize
\item Naive Bayes AUC
\item Naive Bayes ErrRate
\item Naive Bayes Kappa
\item Number Of Binary Features
\item Number Of Classes
\item Number Of Features
\item Number Of Instances
\item Number Of Instances With Missing Values
\item Number Of Missing Values
\item Number Of Numeric Features
\item Number Of Symbolic Features
\item Percentage O fBinary Features
\item Percentage Of Instances With MissingValues
\item Percentage Of MissingValues
\item Percentage Of NumericFeatures
\item Percentage Of SymbolicFeatures
\item Quartile1 Attribute Entropy
\item Quartile1 Mutual Information
\item Quartile2 Attribute Entropy
\item Quartile2 Mutual Information
\item Quartile3 Attribute Entropy
\item Quartile3 Mutual Information
\item REPTreeDepth1AUC
\item REPTreeDepth1ErrRate
\item REPTreeDepth1Kappa
\item REPTreeDepth2AUC
\item REPTreeDepth2ErrRate
\item REPTreeDepth2Kappa
\item REPTreeDepth3AUC
\item REPTreeDepth3ErrRate
\item REPTreeDepth3Kappa
\item RandomTreeDepth1AUC
\item RandomTreeDepth1ErrRate
\item RandomTreeDepth1Kappa
\item RandomTreeDepth2AUC
\item RandomTreeDepth2ErrRate
\item RandomTreeDepth2Kappa
\item RandomTreeDepth3AUC
\item RandomTreeDepth3ErrRate
\item RandomTreeDepth3Kappa
\item StdvNominalAttDistinctValues
\item kNN1NAUC
\item kNN1NErrRate
\item kNN1NKappa    
\end{enumerate}


\end{document}